\documentclass{article}

\usepackage{PRIMEarxiv}

\usepackage[utf8]{inputenc} 
\usepackage[T1]{fontenc}    
\usepackage{hyperref}       
\usepackage{url}            
\usepackage{booktabs}       
\usepackage{amsfonts}       
\usepackage{nicefrac}       
\usepackage{microtype}      
\usepackage{lipsum}
\usepackage{fancyhdr}       
\usepackage{graphicx}       
\usepackage{natbib}

\usepackage{float}
\usepackage{subfigure}

\newtheorem{lemma}{Lemma}

\title{Conformal Prediction for Privacy-Preserving Machine Learning}

\author{
  Alexander~D. Balinsky \thanks{Corresponding author: \texttt{alexander.balinsky22@imperial.ac.uk}} \\ Imperial College London \\
   \And
  Dominik Krzemi\'nski \\
  \And
  Alexander Balinsky \\ Cardiff University
 \\
}

\begin{document}
\maketitle

\begin{abstract}
We investigate the integration of Conformal Prediction (CP) with supervised learning on deterministically encrypted data, aiming to bridge the gap between rigorous uncertainty quantification and privacy-preserving machine learning. Using AES-encrypted variants of the MNIST dataset, we demonstrate that CP methods remain effective even when applied directly in the encrypted domain, owing to the preservation of data exchangeability under fixed-key encryption. We test traditional $p$-value-based against $e$-value-based conformal predictors. Our empirical evaluation reveals that models trained on deterministically encrypted data retain the ability to extract meaningful structure, achieving 36.88\% test accuracy -- significantly above random guessing (9.56\%) observed with per-instance encryption. Moreover, $e$-value-based CP achieves predictive set coverage of over 60\% with 4.3 loss-threshold calibration, correctly capturing the true label in 4888 out of 5000 test cases. In contrast, the $p$-value-based CP yields smaller predictive sets but with reduced coverage accuracy. These findings highlight both the promise and limitations of CP in encrypted data settings and underscore critical trade-offs between prediction set compactness and reliability.
\end{abstract}

\keywords{conformal prediction \and $e$-test statistics \and encryption \and privacy}

\section{Introduction}

Conformal Prediction (CP) offers a principled and robust framework for quantifying prediction uncertainty by generating confidence sets or intervals that, with a user-specified probability, are guaranteed to contain the true outcome \citep{VovkGammShaf2025}. A key strength of CP lies in its {\em distribution-free} guarantees: the coverage probability holds under any data-generating distribution $P$, provided the data are exchangeable. This generality makes CP particularly appealing in real-world applications where distributional assumptions are often violated or unknown.

The core mechanism of CP involves a nonconformity measure, which quantifies the ``strangeness'' or degree of mismatch between a new data point and a reference dataset. This measure is then used to rank candidate outputs and construct prediction sets with calibrated coverage. While traditional CP methods have been extensively studied and applied in various domains (see, e.g., \cite{10.3150/21-BEJ1447, angelopoulos2023conformal} for recent overviews), there has been growing interest in extending CP beyond its classical settings.

CP has been used effectively in areas such as regression, classification, anomaly detection, and time series forecasting \citep{balasubramanian2014conformal}. For example, adaptive conformal methods have been proposed to handle covariate shift \citep{tibshirani2019conformal, Sesia2021}, while split conformal prediction enables scalability to large datasets with strong theoretical guarantees \citep{lei2017distributionfree}. These advances have cemented CP as a versatile tool for trustworthy machine learning, especially in high-stakes environments such as healthcare, finance, and autonomous systems \citep{wisniewski2020application, vazquez2022conformal, lindemann2023safe}.

Recent developments include {\em conformal e-prediction} and the {\em BB-predictor} \citep{pmlr-v230-balinsky24a}, which broaden the applicability of conformal methods. The BB-predictor, in particular, relaxes the standard exchangeability assumption by operating under the weaker condition of cycle invariance, significantly expanding the class of permissible data-generating processes. Furthermore, \citet{GauthierBachJordan2025} demonstrate that $e$-value-based CP can offer tighter control of Type I error and better adaptation in online or adversarial scenarios, where standard CP may be less effective.

In this work, we explore a new frontier for CP: learning under encryption. By extending CP to operate directly on encrypted data, we aim to enable rigorous uncertainty quantification in privacy-preserving machine learning systems.
We define encryption as a mapping from a \textit{message (data) space} to a \textit{ciphertext space}, governed by a \textit{password parameter} (or an encryption key). Employing a consistent password for all data points makes the encryption process {\em deterministic}. This determinism induces a well-defined probability distribution on the ciphertext space and consequently on the set of possible encrypted observations. Given that encryption represents a fixed transformation, it preserves the exchangeability of the underlying data. Therefore, we anticipate that CP will remain effective in this scenario, taking advantage of its inherent distribution-free property. While encryption obscures the original messages, making them seemingly like noise, the deterministic nature of the encryption allows for the potential learning of underlying patterns, as identical data points will consistently encrypt to the same ciphertext.

\section{Methodology}

Our objective is to enable uncertainty quantification in privacy-preserving settings. This introduces a unique challenge: adapting CP to operate effectively on data that remains encrypted throughout the entire machine learning pipeline. To address this, we propose a~two-fold strategy:

\begin{itemize}
\item \textbf{Consistent encryption across all stages:} We ensure that the same deterministic encryption scheme is applied uniformly to all data partitions—training, calibration, and test sets. This consistency is critical to preserving the statistical properties necessary for conformal prediction, particularly exchangeability.

\item \textbf{Nonconformity scoring in the encrypted domain:} All computations, including the derivation of nonconformity scores, are performed directly on encrypted data without decryption. This preserves privacy and tests the feasibility of learning and uncertainty quantification in a fully encrypted setting.
\end{itemize}

CP constructs prediction sets with coverage guarantees derived from nonconformity scores, which measure how unusual a new observation is relative to a reference set. The~central technical challenge in our setting is to compute meaningful nonconformity scores when the data remains in its encrypted form and lacks human-interpretable structure.

For encryption, we employ the widely used Advanced Encryption Standard (AES) with a fixed symmetric key \citep{NISTFIPS197}. This deterministic encryption ensures that identical inputs map to the same ciphertext, enabling consistent model behavior and learnable patterns despite the obfuscation of semantic content.

A simple feedforward neural network is then trained on the deterministically encrypted training set. Importantly, the model never interacts with the plaintext data. Once trained, the model’s outputs (such as loss values) are used to construct nonconformity scores for test instances, thereby enabling the application of conformal prediction techniques in the encrypted domain.

In our application of CP, we investigate two distinct methodologies: the traditional $p$-value approach and the more recent $e$-value approach.

The subsequent lemma serves as the foundational mathematical tool within standard CP theory, drawing upon the statistical concept of the $p$-value from hypothesis testing.

\begin{lemma} \label{CP_lemma}
	Suppose $L_1, \ldots, L_{n+1}$ are exchangeable random variables. Set
	$$
	U = \# \{ i = 1, \ldots, n+1 : \ L_i \geq L_{n+1}\},
	$$
	i.e., the number of $L$ that are at least as large as the last one. Then
	\[
	P \left\{ \frac{U}{n+1} > \epsilon \right\} \geq 1-\epsilon
	\]
	for all $\epsilon\in [0,1]$.
\end{lemma}

The $e$-value approach is based on the result from \cite{pmlr-v230-balinsky24a}:
\begin{lemma} \label{E_lemma}
Suppose $L_1, \ldots, L_{n+1}$ are exchangeable non-negative random variables. Then for any positive $\alpha$:
	
	\begin{equation} \label{main_ineq}
		P \left\{  \frac{L_{n+1}}{\frac{L_1 + \ldots + L_n}{n}}  \geq
		\frac{1}{\alpha} \left(  \frac{1}{1+ \frac{1-1/\alpha}{n}}\right)\right\}
		\leq \alpha.
	\end{equation}
\end{lemma}
This allows to introduce the
\textit{\textbf{BB-predictor}} (bounded from the below predictor):

\begin{equation} \label{bb_ineq}
	P \left\{  L_{n+1} \geq 
	\frac{1}{\alpha} \left(  \frac{1}{1+ \frac{1-1/\alpha}{n}}\right) \times {\frac{L_1 + \ldots + L_n}{n}}  \right\}
	\leq \alpha.
\end{equation}

\

\section{Empirical Results: A Case Study on Encrypted MNIST}
\label{sec:methodology}

We evaluate our approach using the MNIST dataset \citep{lecun1998mnist}, a standard benchmark in image classification, which consists of 60,000 grayscale training images and 10,000 test images of handwritten digits, each of size $28 \times 28$ pixels. Its simplicity and well-understood structure make it a useful testbed for exploring the impact of encryption on both predictive performance and conformal coverage.

To simulate a privacy-preserving setting, we apply deterministic encryption to the entire dataset using the Advanced Encryption Standard (AES) in Cipher Block Chaining (CBC) mode. Specifically, we use a fixed symmetric key (\texttt{key = `abs2kas126oZbdXs'}) and initialization vector (IV; \texttt{iv = `1nsdjah72MdnJ12a'}) to ensure that the same plain data always results in the same ciphered data. This deterministic encryption is crucial to maintaining consistency across training, calibration, and test sets, which in turn preserves the exchangeability assumptions that underlie conformal prediction.

\begin{figure}[t]
	\centering
	\includegraphics[width=0.35\linewidth]{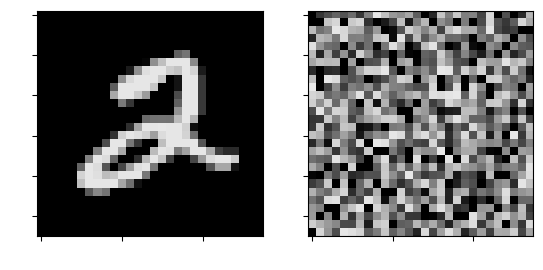}
	\caption{\label{fig:encr_digit} An example sample of a digit from the MNIST dataset (left) and its encrypted version (right).}
\end{figure}

Figure~\ref{fig:encr_digit} visualizes a sample image from the MNIST dataset alongside its encrypted representation. Notably, the resulting ciphertexts are visually uninformative and lack any perceivable structure, making the downstream task of learning and uncertainty quantification particularly challenging.

We train a lightweight feed-forward neural network on the encrypted training images and evaluate its performance using standard classification accuracy and conformal prediction set coverage. Nonconformity scores are derived from the model's softmax outputs, and prediction sets are constructed using split conformal prediction on the encrypted calibration set.

In our results, we examine several key metrics:

\begin{itemize}
\item \textbf{Classification Accuracy:} We assess the performance of models trained on encrypted vs. unencrypted data to quantify the impact of encryption on learnability.
\item \textbf{Coverage and Set Size:} We measure the empirical coverage and average size of the conformal prediction sets to evaluate uncertainty quantification under encryption.
\item \textbf{Robustness to Key Variation:} We conduct ablations to test the sensitivity of the model and CP performance to changes in the encryption key or IV.
\end{itemize}

Preliminary results indicate that while encryption introduces a performance degradation in classification accuracy (as expected), conformal prediction sets maintain valid coverage, highlighting the feasibility of uncertainty quantification even when the input data remains fully encrypted.

\textbf{Code Availability} We make our codebase with AES encryption, model training, and conformal predictions available: \cite{AlexD2025}.

\subsection{Data Visualisation with $t$-SNE}

\begin{figure}[t]
    \centering
    \subfigure[MNIST]{\includegraphics[width=0.32\textwidth]{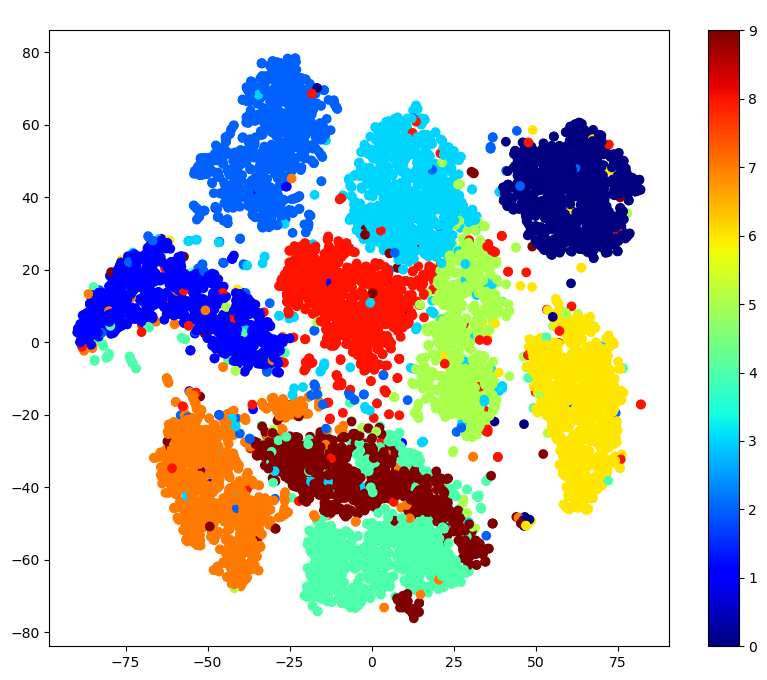}}
    \subfigure[Encrypted MNIST with fixed key]{\includegraphics[width=0.32\textwidth]{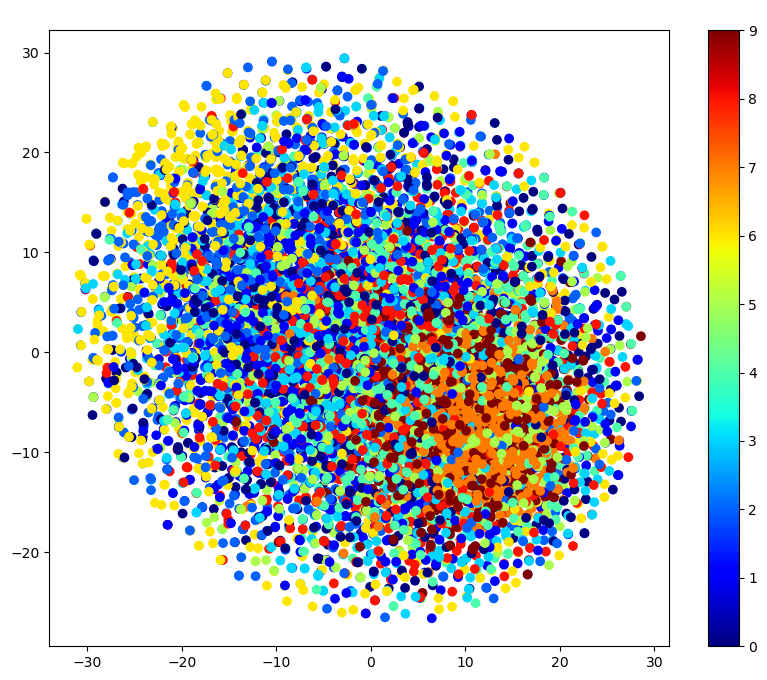}}
    \subfigure[MNIST with randomized key, IV]{\includegraphics[width=0.32\textwidth]{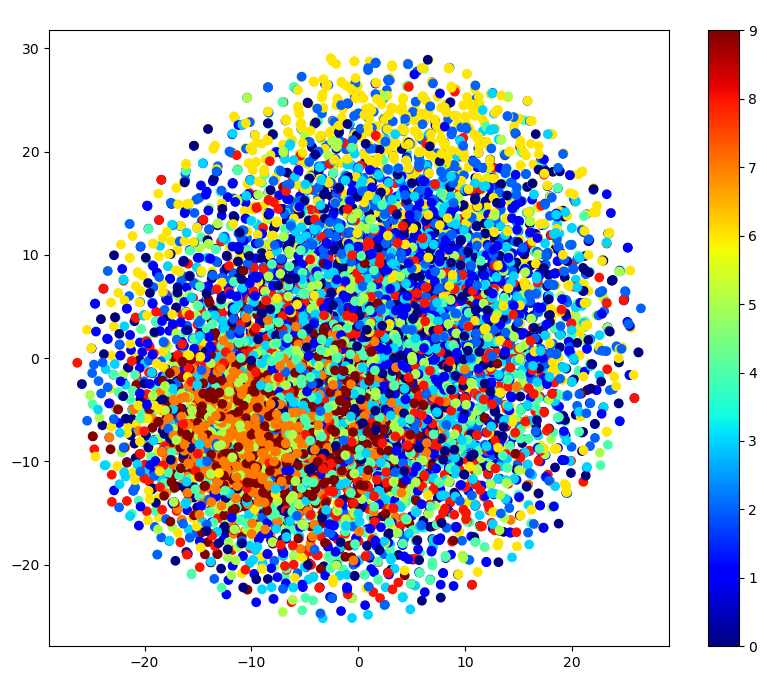}}
    \caption{2-dimensional $t$-SNE projection of the training dataset.}
    \label{fig:tsne_mnist}
\end{figure}

We begin our empirical evaluation with a qualitative analysis of the MNIST dataset and its encrypted variants using dimensionality reduction. Specifically, we compare three scenarios: (a) the original, unencrypted MNIST dataset; (b) the dataset encrypted with a fixed encryption key; and (c) the dataset encrypted with a unique encryption key per image.

To visualize the impact of encryption on data structure, we employ \textit{t-distributed Stochastic Neighbor Embedding} ($t$-SNE) \citep{van2008visualizing}, a widely used non-linear dimensionality reduction technique. $t$-SNE projects high-dimensional data into a lower-dimensional space (typically two or three dimensions), aiming to preserve local similarities and expose underlying structure such as clustering patterns.

Figure~\ref{fig:tsne_mnist}(a) shows the $t$-SNE projection of the first 10,000 samples from the unencrypted MNIST dataset. As expected, the visualization reveals well-separated clusters corresponding to different digit classes, reflecting the strong structure and separability inherent in the original data.

When the dataset is encrypted using a fixed key and initialization vector (see Section~\ref{sec:methodology}), the low-dimensional representation (Figure~\ref{fig:tsne_mnist}(b)) shows a dramatic loss of discernible structure. The clusters largely vanish, suggesting that the encryption process obfuscates the visual and statistical patterns that classifiers typically exploit.

In the most challenging setting, where each image is encrypted with a different key (password), the $t$-SNE projection (Figure~\ref{fig:tsne_mnist}(c)) reveals a near-complete collapse of meaningful structure. The data distribution appears almost random, confirming the intuitive expectation that this level of per-sample encryption maximally disrupts inter-sample relationships.

These visualizations underscore the fundamental difficulty of learning from encrypted data, particularly when conformal prediction techniques, which rely on relative similarity through nonconformity scores, must be applied.

\subsection{Simple Neural Network model}

\begin{figure}[H]
	\centering
    \includegraphics[width=1\linewidth]{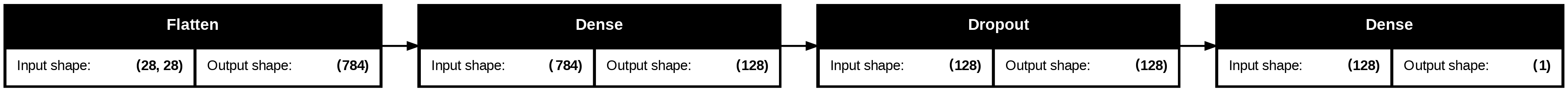}
	\caption{\label{fig:model} The diagram of a simple neural network used. The architecture of the network did not change across the experiments.}
\end{figure}

Due to the disruption of spatial correlations introduced by encryption, conventional Convolutional Neural Networks (CNNs), which rely heavily on such structure, are ill-suited for classification. Consequently, we employ a fully connected feedforward architecture, implemented using the \texttt{TensorFlow Keras} framework (see Figure~\ref{fig:model}). For reproducibility, all experiments were conducted with the random seed set to \texttt{2024}.

We begin by training this model on the original, unencrypted MNIST dataset using hyperparameters: \texttt{batch size} = 64 and 32 epochs. Experiments were repeated 10 times. The model achieves an average training accuracy of 99.35\% and a test accuracy of 98.11\%, confirming that even a relatively simple architecture is sufficient to capture the underlying structure of the dataset.

We then assess the same model architecture under encryption. When trained on MNIST images encrypted with a fixed key and initialization vector (AES encryption; see Section~\ref{sec:methodology}), the model attains an average training accuracy of 39.48\% and a test accuracy of 36.88\%. Although substantially degraded relative to the unencrypted case, this performance still exceeds the 10\% baseline expected from random guessing in a 10-class setting. This suggests that, despite the loss of visual interpretability, some latent patterns remain accessible to the network under deterministic encryption. In contrast, training the same model on the MNIST dataset with randomized encryption per sample (a unique key per image) results in a test accuracy of 9.56\%, indistinguishable from random guessing. This outcome aligns with our expectations, as the individualized encryption destroys any consistent mapping between data instances, rendering the learning task infeasible under this setup.

\subsection{Encrypted data and Inductive Conformal Prediction}

Following the training of our model on the encrypted dataset, we discard the original training data and retain only the trained model and its associated loss function for computing nonconformity scores.

\begin{figure}[t]
	\centering
    \includegraphics[width=0.41\linewidth]{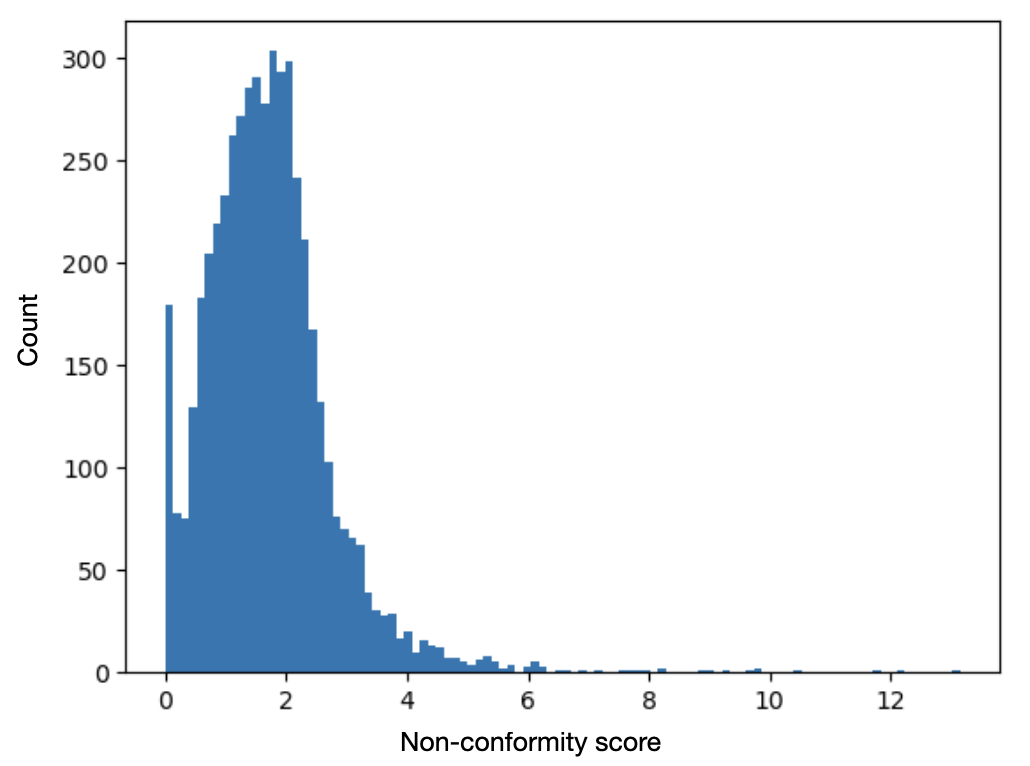}
	\includegraphics[width=0.4\linewidth]{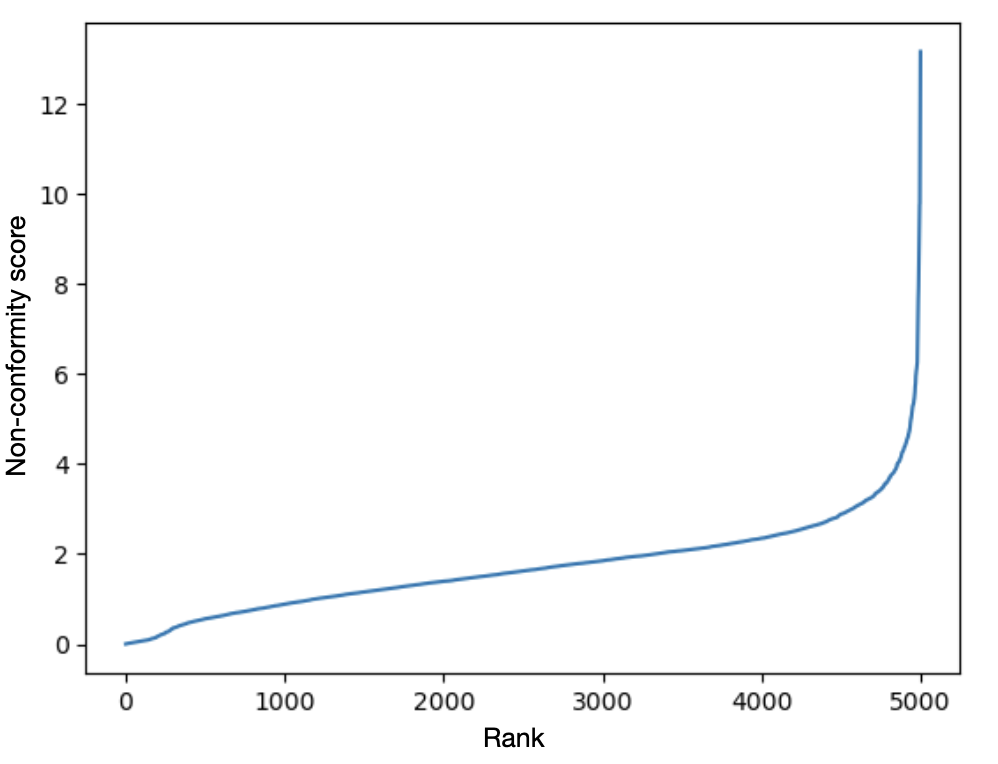}
	\caption{\label{fig:calibr_values} The histogram and the sorted loss function values for the  {\em CalibrationSet}.}
\end{figure}

Our focus now shifts to the \textit{test set}, which is randomly partitioned into two disjoint subsets of equal size: the {\em Calibration Set} and the {\em Conformal Prediction Test Set}. The Calibration Set is used exclusively to estimate the quantiles of the nonconformity score distribution, a critical step in constructing valid conformal prediction sets. Figure~\ref{fig:calibr_values} visualizes the empirical distribution of nonconformity scores on the Calibration Set, including both a histogram and the sorted values, offering insight into the model's confidence under encryption.

The baseline classification accuracy of the model on encrypted test data is 36.88\%, as previously noted. However, in the context of conformal prediction, our primary concern is coverage, i.e., the frequency with which the true label is contained in the prediction set. While traditional conformal prediction frameworks often aim for high coverage thresholds (e.g., 95\%), such targets are impractical in our encrypted setting. At 95\% coverage, the resulting prediction sets frequently include all ten possible digit classes, offering limited discriminative value.

Instead, we evaluate the system at a more pragmatic coverage level of 60\%, which allows for narrower and more informative prediction sets while still offering meaningful uncertainty quantification. This adjusted target better reflects the trade-offs required when applying conformal methods to encrypted domains with reduced model fidelity.

\subsubsection{Inductive Conformal Prediction with e-test statistics}

Given a miscoverage level of $\alpha = 0.4$ and a test set size of $n = 5000$, the \textit{BB-predictor} inequality (Eq.~\ref{bb_ineq}) ensures that for any test pair $(x, y)$ with $x$ drawn from the {\em ConformalPredictionTestSet}, the following bound holds:
$$
P\{ LossFunction \geq  4.29327 \} < 0.4.
$$
Applying this threshold across the entire {\em ConformalPredictionTestSet}, we construct prediction sets accordingly.

\begin{table}[h!]
\centering
\caption{Label set size distribution ($e$-test).}

\begin{tabular}{ |l|c|c|c|c|c|c|c|c|c|c|c| } 
 \hline
 Size of label set & 0 & 1 & 2 & 3 & 4 & 5 & 6 & 7 & 8 & 9 & 10\\ 
 \hline
Number of examples & 0 & 17 & 236 & 171 & 139 & 199 & 401 & 1705 & 1694 & 428 & 10\\
 \hline
\end{tabular}
\end{table}

The central question is: \textit{how often does the true label fall within the prediction set}? In our experiment, we observe that 4888 out of 5000 prediction sets include the true label, corresponding to a realized coverage of approximately 97.76\%.

\subsubsection{Inductive Conformal Prediction with p-value statistics}

For comparison, we now apply the standard $p$-value CP framework. According to Lemma~\ref{CP_lemma}, with $\epsilon = 0.4$ and $n = 5000$, the appropriate quantile for thresholding is given by the $(1 - \epsilon)(n + 1)$-th order statistic of the calibration losses, which corresponds to index $3000$. The resulting threshold is 1.8497.

Applying this threshold to the same {\em ConformalPredictionTestSet}, we compute prediction sets under the CP framework. This yields a total of 2965 prediction sets that successfully contain the true label, achieving a realized coverage of 59.3\%.

\begin{table}[h!]
\centering
\caption{Label set size distribution ($p$-test).}
\begin{tabular}{ |l|c|c|c|c|c|c|c|c|c|c|c| } 
 \hline
 Size of label set & 0 & 1 & 2 & 3 & 4 & 5 & 6 & 7 & 8 & 9 & 10\\ 
 \hline
Number of examples & 5 & 842 & 3017 & 956 & 177 & 3 & 0 & 0 & 0 & 0 & 0\\
 \hline
\end{tabular}
\end{table}

\noindent These results highlight a notable trade-off: while the CP method produces significantly smaller prediction sets, it does so at the cost of reduced empirical coverage. In contrast, the BB-predictor offers a looser guarantee, resulting in higher observed coverage, even in the presence of encrypted and structurally obfuscated data.

\section{Conclusion}

In this work, we investigated the integration of conformal prediction with machine learning models operating on encrypted data. Using the AES-encrypted MNIST dataset as a testbed, we demonstrated the practical feasibility of applying conformal prediction frameworks under deterministic encryption, and highlighted the nuanced behavior of nonconformity scores in this setting.

Our empirical findings reveal that, despite encryption disrupting spatial and statistical structure, simple neural models are still capable of extracting learnable patterns from encrypted inputs. However, we also observed that conformal prediction sets tend to grow in size under encryption, particularly when aiming for high coverage guarantees. This emphasizes a key trade-off: as encryption obfuscates structure, maintaining valid coverage often necessitates less informative (i.e., larger) prediction sets.

These observations motivate several important avenues for future work. First, there is a clear need for the development of novel nonconformity scoring functions tailored to the encrypted domain, which may exploit the deterministic properties of the encryption or the invariants preserved across examples. Second, generalizing this approach to broader encryption schemes (e.g., homomorphic or probabilistic encryption) and testing more complex datasets will be essential for assessing real-world applicability. Ultimately, our results lay the foundation for principled uncertainty estimation in privacy-first machine learning pipelines, opening the door to secure yet interpretable predictive systems.

\bibliographystyle{plainnat}

\begin{thebibliography}{16}
	\providecommand{\natexlab}[1]{#1}
	\providecommand{\url}[1]{\texttt{#1}}
	\expandafter\ifx\csname urlstyle\endcsname\relax
	\providecommand{\doi}[1]{doi: #1}\else
	\providecommand{\doi}{doi: \begingroup \urlstyle{rm}\Url}\fi
	
	\bibitem[Angelopoulos and Bates(2023)]{angelopoulos2023conformal}
	A.N. Angelopoulos and S.~Bates.
	\newblock \emph{Conformal Prediction: A Gentle Introduction}.
	\newblock Foundations and Trends in Machine Learning Series. Now Publishers,
	2023.
	\newblock ISBN 9781638281580.
	\newblock URL \url{https://books.google.co.uk/books?id=3gK8zwEACAAJ}.
	
	\bibitem[Balasubramanian et~al.(2014)Balasubramanian, Ho, and
	Vovk]{balasubramanian2014conformal}
	Vineeth Balasubramanian, Shen-Shyang Ho, and Vladimir Vovk.
	\newblock \emph{Conformal prediction for reliable machine learning: theory,
		adaptations and applications}.
	\newblock Newnes, 2014.
	
	\bibitem[Balinsky and Balinsky(2024)]{pmlr-v230-balinsky24a}
	A.~A. Balinsky and A.~D. Balinsky.
	\newblock Enhancing conformal prediction using e-test statistics.
	\newblock In Simone Vantini, Matteo Fontana, Aldo Solari, Henrik Boström, and
	Lars Carlsson, editors, \emph{Proceedings of the Thirteenth Symposium on
		Conformal and Probabilistic Prediction with Applications}, volume 230 of
	\emph{Proceedings of Machine Learning Research}, pages 65--72. PMLR, 09--11
	Sep 2024.
	\newblock URL \url{https://proceedings.mlr.press/v230/balinsky24a.html}.
	
	\bibitem[Balinsky(2025)]{AlexD2025}
	A.~D. Balinsky.
	\newblock Encryption\uppercase{C}onformal\uppercase{COPA}2025 github
	repository.
	\newblock
	\url{https://github.com/AlexanderBalinsky/EncryptionConformalCOPA2025/},
	2025.
	
	\bibitem[Bates et~al.(2021)Bates, Cand{\`e}s, Lei, Romano, and
	Sesia]{Sesia2021}
	Stephen Bates, Emmanuel~J. Cand{\`e}s, Lihua Lei, Yaniv Romano, and Matteo
	Sesia.
	\newblock Testing for outliers with conformal p-values.
	\newblock \emph{The Annals of Statistics}, 2021.
	\newblock URL \url{https://api.semanticscholar.org/CorpusID:233301258}.
	
	\bibitem[Fontana et~al.(2023)Fontana, Zeni, and Vantini]{10.3150/21-BEJ1447}
	M.~Fontana, G.~Zeni, and S.~Vantini.
	\newblock {Conformal prediction: A unified review of theory and new
		challenges}.
	\newblock \emph{Bernoulli}, 29\penalty0 (1):\penalty0 1 -- 23, 2023.
	\newblock \doi{10.3150/21-BEJ1447}.
	\newblock URL \url{https://doi.org/10.3150/21-BEJ1447}.
	
	\bibitem[Gauthier et~al.(2025)Gauthier, Bach, and
	Jordan]{GauthierBachJordan2025}
	E.~Gauthier, F.~Bach, and M.~Jordan.
	\newblock E-\uppercase{V}alue \uppercase{E}xpand the \uppercase{S}cope of
	\uppercase{C}onformal \uppercase{P}rediction.
	\newblock Technical report, 2025.
	\newblock hal-05003357.
	
	\bibitem[LeCun(1998)]{lecun1998mnist}
	Yann LeCun.
	\newblock The {MNIST} database of handwritten digits.
	\newblock \emph{\url{http://yann.lecun.com/exdb/mnist/}}, 1998.
	
	\bibitem[Lei et~al.(2017)Lei, G'Sell, Rinaldo, Tibshirani, and
	Wasserman]{lei2017distributionfree}
	Jing Lei, Max G'Sell, Alessandro Rinaldo, Ryan~J. Tibshirani, and Larry
	Wasserman.
	\newblock Distribution-free predictive inference for regression, 2017.
	\newblock URL \url{https://arxiv.org/abs/1604.04173}.
	
	\bibitem[Lindemann et~al.(2023)Lindemann, Cleaveland, Shim, and
	Pappas]{lindemann2023safe}
	Lars Lindemann, Matthew Cleaveland, Gihyun Shim, and George~J Pappas.
	\newblock Safe planning in dynamic environments using conformal prediction.
	\newblock \emph{IEEE Robotics and Automation Letters}, 8\penalty0 (8):\penalty0
	5116--5123, 2023.
	
	\bibitem[NIST(2001)]{NISTFIPS197}
	NIST.
	\newblock Advanced {E}ncryption {S}tandard ({AES}).
	\newblock Federal Information Processing Standards Publication FIPS 197-upd1,
	U.S. Department of Commerce, 2001.
	\newblock URL \url{https://doi.org/10.6028/NIST.FIPS.197-upd1}.
	\newblock Updated May 9, 2023.
	
	\bibitem[Tibshirani et~al.(2020)Tibshirani, Barber, Candes, and
	Ramdas]{tibshirani2019conformal}
	Ryan~J. Tibshirani, Rina~Foygel Barber, Emmanuel~J. Candes, and Aaditya Ramdas.
	\newblock Conformal prediction under covariate shift, 2020.
	\newblock URL \url{https://arxiv.org/abs/1904.06019}.
	
	\bibitem[Van~der Maaten and Hinton(2008)]{van2008visualizing}
	Laurens Van~der Maaten and Geoffrey Hinton.
	\newblock Visualizing data using t-{SNE}.
	\newblock \emph{Journal of machine learning research}, 9\penalty0 (11), 2008.
	
	\bibitem[Vazquez and Facelli(2022)]{vazquez2022conformal}
	Janette Vazquez and Julio~C Facelli.
	\newblock Conformal prediction in clinical medical sciences.
	\newblock \emph{Journal of Healthcare Informatics Research}, 6\penalty0
	(3):\penalty0 241--252, 2022.
	
	\bibitem[Vovk et~al.(2005)Vovk, Gammerman, and Shafer]{VovkGammShaf2025}
	V.~Vovk, A.~Gammerman, and G.~Shafer.
	\newblock \emph{Algorithmic Learning in a Random World}.
	\newblock Springer-Verlag, Berlin, Heidelberg, 2005.
	\newblock ISBN 0387001522.
	
	\bibitem[Wisniewski et~al.(2020)Wisniewski, Lindsay, and
	Lindsay]{wisniewski2020application}
	Wojciech Wisniewski, David Lindsay, and Sian Lindsay.
	\newblock Application of conformal prediction interval estimations to market
	makers’ net positions.
	\newblock In \emph{Conformal and probabilistic prediction and applications},
	pages 285--301. PMLR, 2020.
	
\end{thebibliography}

\end{document}